\pdfoutput=1
\documentclass{article}
\usepackage{spconf,amsmath,float,graphicx,subcaption, hyperref,pdfpages}

\newcommand\norm[1]{{\lVert#1\rVert}_{2}}
\newcommand\normone[1]{{\lVert#1\rVert}_{1}}
\usepackage{amsfonts}

\title{{F\MakeLowercase{ew}GAN}: Generating from the Joint Distribution of a Few Images}
\name{Lior Ben-Moshe$^{1}$ \qquad Sagie Benaim$^{2}$ \qquad Lior Wolf$^{1}$}
  
\address{$^{1}$Tel Aviv University,  $^{2}$University of Copenhagen}
\begin{document}
%\ninept
%
\maketitle
\begin{abstract}
We introduce FewGAN, a generative model for generating novel, high-quality and diverse images whose patch distribution lies in the joint patch distribution of a small number of $N>1$ training samples. 
% The model is trained to capture the internal patch distribution of multiple images, and is able to generate novel, high quality, diverse samples whose patches lie withing the internal patch distribution of the training samples. 
The method is, in essence, a hierarchical patch-GAN that applies quantization at the first coarse scale, in a similar fashion to VQ-GAN, followed by a pyramid of residual fully convolutional GANs at finer scales. 
Our key idea is to first use quantization to learn a fixed set of patch embeddings for training images. We then use a separate set of side images to model the structure of generated images using an autoregressive model trained on the learned patch embeddings of training images. 
Using quantization at the coarsest scale allows the model to generate both conditional and unconditional novel images. 
% , by using an autoregressive prior. 
Subsequently, a patch-GAN renders the fine details, resulting in high-quality images. In an extensive set of experiments, it is shown that FewGAN outperforms baselines both quantitatively and qualitatively. 
\end{abstract}


\section{Introduction}
\label{sec:intro}

We present a generative model for   
% the first work, as far as we can ascertain,  that is
% able 
generating novel, high-quality, and diverse samples whose patch distribution lies in the joint patch distribution of a small number of training samples, in a coherent manner. Our method is capable of generating both conditional and unconditional images, as well as solving a variety of image manipulation tasks, including editing, inpainting and harmonization. In all of these cases, our model produces high-quality results that preserve the internal patch statistics of the training images. All tasks are achieved with the same generative network, without any further training. 

Generating a coherent and realistic image, in an unsupervised manner, from the joint patch distribution of a small number of $N>1$ samples is a challenging problem. First, the joint patch distribution is not evident in any of the training images, yet we are interested in generating a realistic image that depicts patches from multiple images simultaneously in the same image. Second, one has to avoid both a mode collapse in which all patches are taken from the same input image and the mode collapse of generating an image that is very similar to a single training image. 

As a main tool for addressing these two issues, 
we model the patch distribution of training images 
% of training images 
% in a side information of unlabeled images, taken from random sources (naturally, there is no shortage of such images). 
% These images are encoded 
using a vector-quantized (VQ) basis learned on the $N$ training images. We then use a side dataset of unlabeled images, taken from random sources (naturally, there is no shortage of such images)
% . and the distribution observed on the side-dataset is used 
to train an autoregressive model that is able to combine patches that have originated from the training images, but preserve the structure of images from the side dataset. 

\section{Related Work}
\label{sec:related}
In typical unconditional image generation, the learner is provided with a large collection of images and is asked to model the underlying distribution of those images as well as generate novel samples from this distribution~\cite{gan, stylegan2, tamingtransformers, swappingae}.
At the other end of the scale, recent methods model the internal distribution of patches of a single image, thus enabling the generation of novel samples that depict the same internal statistics as the single image~\cite{singan, hpvaegan, gpnn}. 
In this work, we are interested in generating images from the joint patch distribution of a small number of samples $N > 1$. This is different from recent few-shot generation setting, which attempt to model the external distribution of a new domain given only a few images at training/fine-tuning (e.g. faces to emoji, photo to sketch, etc.) \cite{stylegan-nada,li2020fewshot, zhao2020differentiable, liu2020towards}. Instead, we are interested in modeling the internal joint patch distribution of all $N$ images and generating novel samples that capture these patches. As shown in Sec.~\ref{sec:results}, other models are either unable to learn with such small number of samples (mode collapse), or are unable to generate realistic images, as they place patches from different images in an unrealistic manner.

\section{Method}
\label{sec:method}

% We first describe our novel training scheme which incorporates reference samples in order to guide the generator during training.
% We then outline our novel hierarchical patch GAN,  which is inspired by VQ-GAN~\cite{tamingtransformers} at the coarsest scale, and is followed by a pyramid of residual fully convolutional GANs at finer scales.

% \subsection{Training}
% \label{ssec:training}
% Since our method is fully unsupervised, there is unlimited supply of images available online which can guide the generator during its learning process. Therefore, the source of randomness during the generator's training comes from a reference dataset, instead of random noise or VAE \cite{vae}. Note that the size of the reference set must be much larger than the size of the training set, which is small by definition.

%We will hereby note $x_{train}$ as an image from the training set, and $x_{ref}$ as an image from the reference set. 

% \subsection{Multi-Scale architecture}
% \label{ssec:multi-scale}
We employ $T + 1$ scales, where $0$ is the coarsest scale, and $T$ is the finest scale. The role of the first scale is to generate structural diversity. The generator at scale $t = 1, \dots, T$ is trained to provide a residual signal. Thus, each generator progressively adds detail to the upscaled version obtained from the previous generator.  

% \subsubsection{VQ-GAN}
%\subsubsection{Quantization based layer}
%\label{sssec:vq-gan}

The coarsest scale $0$ consists of a fully convolutional encoder $E$ and a decoder $Dec$. $E$ encodes image patches to a fixed set of codes from a learned, discrete codebook $\mathbf{Z} = \{z_k\}^{K}_{k=1} \subset R^{n_z}$. $K$ is the number of codebook entries and $n_z$ is the dimensionality of codes. 

For a given image $x_0 \in \mathbb{R}^{H \times W \times 3}$, the generation at this scale is purely conditioned:
\begin{equation}
    \hat{x}_0 = G_0(x_0) = Dec(Z(E(x_0))) 
\end{equation}
$E$ maps $x_0$ into a latent space code of shape $z \in \mathbb{R}^{h \times w \times n_{z}}$. $z$ can be viewed as the encoding of $h \times w$ patches of dimension $n_{z}$. Each such patch encoding, $z^{ij} \in \mathbb{R}^{n_z}$, is quantized using $Z$, the quantization operator, onto its closest codebook entry ${z_k}$, creating a quantized version of $z$, $z_q$. Next, $Dec$ generates $\hat{x}_0 \in \mathbb{R}^{H \times W \times 3}$ conditioned on $z_q$. For ease of notation we denote $Dec(Z(E(\cdot)))$ as $G_0(\cdot)$.

In order to learn a context-rich vocabulary of the joint patch distribution of the training set, we apply adversarial loss (Eq.~\ref{eq-adv}) and reconstruction loss (Eq.~\ref{eq-vq}) to the training samples:
\begin{align}
\label{eq-adv}
%\begin{split}
     \mathcal{L}_{{Adv}_0}\{G_0,D_0\}(x_0)  & = \min_{G_0} \max_{D_0}( \mathbf{E}[D_{0}(x_0)]  -\mathbf{E}[D_{0}(\hat{x}_0)] \nonumber \\   & -    \lambda_{gp}\mathbf{E}[(\norm{{\nabla}_{\overline{x}_0}  D_{0}(\overline{x}_0)-1})^2])
%\end{split}
\end{align}
The adversarial loss is given by the WGAN-GP \cite{wgan-gp} loss, where $\mathbf{E}$ is the mean over $D_0$’s output, $\overline{x}_0 = \varepsilon x_0 + (1-\varepsilon)\hat{x}_0$, for $\varepsilon$ sampled uniformly between 0 and 1, and $\lambda_{gp}$ is the gradient penalty weight. The VQ loss term is given by:
\begin{equation}
\label{eq-vq}
\begin{split}
    & \mathcal{L}_{VQ}\{G_0\}(x_0)  = \norm{x_0-\hat{x}_0}^2 \\ & + \norm{sg[E(x_0)]-z_q}^2  + \beta \norm{E(x_0)-sg(z_q)}^2
\end{split}
\end{equation}
where $sg$ denotes the stop-gradient operation, which passes zero gradient during backpropagation. The first term is the reconstruction loss, the second term is the codebook loss, and the third term is the commitment loss, with $\beta$ set to $0.25$, as in VQ-VAE~\cite{vqvae}.

%Since the number of codebook entries $K$ is finite, codebook entries are encouraged to capture only the joint patch distribution of the training set. This happens because the discriminator is trained using the training set only, so using patches from a side-dataset will cause a high penalty from the discriminator. 
%Therefore, out-of-distribution images, will be encoded into entries in the codebook that resemble the closest patches of images in the training set. This forces out-of-distribution images to be decoded using patches from the training set.

In order to generate images of a diverse structure, we use a side-dataset, which is separate from the training set. This is used as a source of randomness for the generator and later as a training set for the auto-regressive generator. To keep the structure of an external image, we add $\mathcal{L}_{SSIM}\{G_0\}(s_0, \hat{s}_0)$ \cite{ssim} loss between the generated image and the external image. Since we want to keep only the structure, while generating content only from the joint patch distribution of the training set, we apply adversarial loss only for the generator, i.e. the discriminator does not see any image from the side-dataset as real input. 

Denote by $s_0 \in \mathbb{R}^{H \times W \times 3}$ an image from the side-dataset in the coarsest scale. The loss is as follows:
\begin{equation}
\label{eq-adv-ref}
    \mathcal{L}_{{Adv-ref}_0}\{G_0\}(s_0)= -\mathbf{E}[D_{0}(\hat{s}_0)]
\end{equation}
where $\hat{s}_0 = G_0(s_0)$ and $\mathbf{E}$ is the mean over the output of $D_0$.

We also wish to encourage the model to learn a realistic mapping between patches, e.g., prevent it from mapping water as sky. To enforce this, we concatenate pixel coordinates to the encoded vector before the vector quantization layer. In this manner, each codebook entry also has positional encoding embedded into it. The positional encoding pixel coordinates are defined as follows for each pair of pixel coordinates (i,j): $i' = \frac{2i}{W-1}-1$ and $j' = \frac{2j}{H-1}-1$, uniformly mapped to the range $[{-1}, 1]$. Since the dimensionality of the code vector $n_z$ is usually much larger than 2 coordinates $(i',j')$, we concatenate them $\lambda_{pos}$ times, where $\lambda_{pos}$ is a  hyperparameter.

In addition, since the coarsest scale sets the structure for the rest of the scales, we want to make sure that the structure is semantically continuous, i.e. that adjacent patches, of the same semantic object, will be from the patch distribution of the same image (e.g. when generating sky, use the same distribution of a specific image). To enforce this, we apply spatial continuity loss, originally introduced in \cite{continuity}, to the encoder's output (for both $x_0$ and $s_0$):
\begin{equation}
\label{eq-continuity}
    \mathcal{L}_{Continuity}\{E\}(x_0,s_0) = \mathcal{L}_{Con}(E(x_0)) + \mathcal{L}_{Con}(E(s_0)) \nonumber
\end{equation}
where the spatial continuity loss $\mathcal{L}_{Con}$ is defined as follows:
\begin{equation}
    \mathcal{L}_{Con}(m) = \sum_{i=1}^{W-1} \sum_{j=1}^{H-1} \normone{m_{i+1,j}-m_{i,j}} + \normone{m_{i,j+1}-m_{i,j}} \nonumber
\end{equation}
where $W$ and $H$ represent the width and height of an input image, while $m_{i,j}$ represents the pixel value at $(i, j)$ in the encoded output $m=E(\cdot)$.

%We consider the patch distribution of images $x_1, \dots, x_k$. For a given image $x_i$, and for a given scale $n=0,1,2,\dots,N$, 
%we consider the distribution of image of the same patch distribution as $x_i$ at this scale to be $P^n_{x_i}$. In other words, since each image is composed of many patches at different scales, we learn from image $x_i$, a distribution of patches $P^n_{x_i}$, that shares the same scale-dependent patch distributions. 
%The downsampling factor of each scale $n$ is given by $r^{N-n}$ for some $r > 1$. Scale $N$, therefore, is the original image resolution.

%At coarse scale 0, we model the patch distribution of images using a patch-based Vector Quantised GAN, a variant of the original Vector Quantised VAE, which uses discriminator loss in addition to reconstruction loss. 

%\subsubsection{Residual hierarchical patch-GAN}
%\label{sssec:res-patch-gan}
Following the first scale, in scales 1 to T, our method employs a patch-GAN~\cite{patchgan} for each scale, using a generator ${G_t}$ and discriminator ${D_t}$. As stated, ${G_t}$ is trained in a residual manner, learning to add details to samples from the previous scale. 
For $t>0$, let $\hat{x}_{t-1}$ be the output of the previous scale, $\uparrow \hat{x}_{t-1}$ be the result of upsampling $\hat{x}_{t-1}$ to the scale of level $t$, and $x_t$ the real input image for scale $t$. We define $\hat{x}_t$ to be:
\begin{equation}
    \hat{x}_t = \uparrow \hat{x}_{t-1} + G_{t}(\uparrow \hat{x}_{t-1})
\end{equation}
${D_t}$ produces a single-channel activation map of the same dimension as its input, indicating whether each patch of the input is real or fake, based on the effective receptive field $r$. 

When using a patch-GAN, we follow a similar procedure to SinGAN~\cite{singan} and HP-VAE-GAN~\cite{hpvaegan}, using a fully convolutional generator and a discriminator of a fixed effective receptive field $r$, while varying the resolution of $x$ at each scale. In scales 1 to T, the receptive fields become smaller, and the top-level generators introduce fine textural details. At these scales, we wish to encourage quality over diversity, which patch-GAN \cite{patchgan} does well.

When training each scale $t>0$, only $G_t$ and $D_t$ are trained, while $G_0$, $\dots$, $G_{t-1}$ are frozen. The loss used is:
\begin{align}
\min_{G_t} \max_{D_t} \mathcal{L}_t \{G_t, D_t\} & = \mathcal{L}_{{Adv}_t}\{G_t, D_t\} + \mathcal{L}_{{Adv-ref}_t}\{G_t\} \nonumber \\  + \mathcal{L}_{SSIM}\{G_t\} &+ \mathcal{L}_{Reconstruction}\{G_t\}    
\end{align}
such that $\mathcal{L}_{{Adv}_t}\{G_t, D_t\}$, $\mathcal{L}_{{Adv-ref}_t}\{G_t\}$ and $\mathcal{L}_{SSIM}\{G_t\}$ are defined the same as for scale 0, but with $G_t$ (resp. $D_t$) instead of $G_0$ (resp. $D_0$). $\mathcal{L}_{Reconstruction}$ is defined as $\norm{x_t-\hat{x}_t}^2$.

% \begin{equation}
% \begin{split}
%     \mathcal{L}_{{Adv}_n}\{G_n,D_n\}(x_n) & =  \min_{G_n} \max_{D_n}( \mathbf{E}[D_{n}(x_n)] \\ & -\mathbf{E}[D_{n}(\hat{x}_n)] \\ & -  \lambda_{gp}\mathbf{E}[(\norm{{\nabla}_{\overline{x}_n} D_{n}(\overline{x}_n)-1})^2])
% \end{split}
% \end{equation}
% \begin{equation}
%     \mathcal{L}_{{Adv-ref}_n}\{G_n\}(s_n)= -\mathbf{E}[D_{n}(\hat{s}_n)]
% \end{equation}

% \begin{equation}
%     \mathcal{L}_{SSIM}{G_n}(s_n, \hat{s}_n)
% \end{equation}

% \begin{equation}
%     \mathcal{L}_{Reconstruction}\{G_n\}(x_n)= \norm{x_n-\hat{x}_n}^2
% \end{equation}

% \subsection{Loss Term}
% \label{ssec:loss}

% For scale 0:
% \begin{equation}
% \begin{split}
% \min_{G_n} \max_{D_n} \mathcal{L}_0\{E,Z,G_0,D_0\} & = \mathcal{L}_{{Adv}_0}\{E,Z,G_0,D_0\} \\ & + \mathcal{L}_{{Adv-ref}_0}\{E,Z,G_0\} \\ & + \mathcal{L}_{SSIM}\{E,Z,G_0\} \\ & + \mathcal{L}_{VQ}\{E,Z,G_0\} \\ & + \mathcal{L}_{Continuity}\{E,Z,G_0\}
% \end{split}
% \end{equation}

% For scale $n>0$:
% \begin{equation}
% \begin{split}
% \min_{G_n} \max_{D_n} \mathcal{L}_n \{G_n, D_n\} & = \mathcal{L}_{{Adv}_n}\{G_n, D_n\} + \mathcal{L}_{{Adv-ref}_n}\{G_n\} \\ & + \mathcal{L}_{SSIM}\{G_n\} + \mathcal{L}_{Reconstruction}\{G_n\}    
% \end{split}
% \end{equation}

%\subsection{Auto-regressive generator}
With $E$ and $Z$ available, we can
now represent the side-dataset images in terms of the codebook indices of
their encodings. For a given image $s \in \mathbb{R}^{H \times W \times 3}$, we consider $c = Z(E(s))\in \{0, \dots ,K-1 \}^{H*W}$. Image generation can then be formulated as an auto-regressive prediction: Given indices $c_{<i}$, a PixelCNN model learns to predict the distribution of the next location, i.e. $\mathcal{P}(c_i
|c_{<i})$. %to compute the likelihood of the full representation as $\mathcal{P}(c) = \prod_{i=0}^{H*W} \mathcal{P}(c_i|c_{<i})$. This allows us to 
The training procedure directly maximizes the log-likelihood with respect to this autoregressive processing of the images $s$ in the side dataset. % side-dataset representations: 
%\begin{equation}
%    \mathcal{L}_{PixelCNN} = \mathbf{E}[-\log(\mathcal{P}(c))]
%\end{equation}

\noindent{\bf Inference\quad } Once we have the PixelCNN trained, we can generate $c$ via ancestral sampling, and then use $Dec$, $G_1,$ $\dots,$ $G_T$ to unconditionally generate novel images from the joint patch distribution of the training images.

Using FewGAN for image manipulation tasks, such as editing and harmonization, is straightforward by applying $G_0, \dots, G_T$ on the edited input. As for multi-modal inpainting, the multi-modality is derived from the PixelCNN model in the following manner: (1) mask the corresponding pixels of the edited input in the discrete latent space, i.e. after applying $E$ and $Z$. (2) PixelCNN fills the occluded section. (3) apply $Dec, G_1, \dots, G_T$ on the output of PixelCNN.

% \subsection{Inference}
% \label{ssec:prior}
% At inference time, to generate novel sampels, we train auto-regressive generator, such as PixelCNN, over the vq-gan discrete indices, and then unconditionally generate novel, high quality, diverse images.

% Since we already have the reference set available, after training the generator, we encode all the reference set by using the quantization layer, and use the generated indices as a training set for the PixelCNN. At the end of this process we get a single, end-to-end model that handles both unconditional and conditional generation. 

\section{Results}
\label{sec:results}
To the best of our knowledge, our work is the first that enables the generation of novel images from the joint patch distribution of a small number of training samples. We have tested a number of methods that require a large collection of images and they resulted in a mode collapse~\cite{tamingtransformers, swappingae, stylegan2, vqvae}. Methods dedicated to handling relatively small datasets \cite{stylegan2_ada, fastgan, dataefficientgan} also result in mode collapse when using only few images. 

As for single-image generation methods, SinGAN's \cite{singan} extension to multiple images was not successful. On the other hand, we were able to extend GPNN~\cite{gpnn}, the nearest neighbor version of SinGAN \cite{singan}, to generate images conditionally, by using patches from multiple images. Since this method lacks the ability of unconditional generation, we only show its result in the qualitative section. We do not include it in the quantitative evaluation, since it is based on unconditional evaluation only. We were able to successfully extend HP-VAE-GAN~\cite{hpvaegan} to work with multiple images; therefore, it will be used as our primary baseline. The extension was straightforward by simply passing multiple images into the VAE \cite{vae} in the coarsest scale, thus enabling both conditional and unconditional generation.

Since the number of images $N$ is quite small, we have also tried to concatenate all the training images into a single image, and to train a single-image generative model on that image. We used SinGAN \cite{singan} as the single-image generative model, and call this baseline Concat-SinGAN. The results of this method are relatively poor and unrealistic, so we only show the quantitative results of this baseline.

%\subsection{Quantitative Evaluation}
%\label{ssec:quantitative-eval}

To evaluate our method quantitatively, we evaluate the realism and diversity of the generated samples. For realism, we use the KID~\cite{kid} measure. FID~\cite{fid} is also reported, despite being shown to be unreliable for small datasets, due to its bias towards the dataset size $N$ \cite{kid, stylegan2_ada, fid_bias}. To further evaluate realism, we conducted a user study. Our study involved 25 users and 15 images generated by our method, the primary baseline (HP-VAE-GAN \cite{hpvaegan}) and Concat-SinGAN. For each image, users were asked to rank how real the generated image looks on a scale of 1 to 5. For diversity, we use the measure introduced in SinGAN \cite{singan}, which computes the average standard deviation over all pixel values along the channel axis of 200 generated images.

% in two aspects: (1) evaluate that the model is generating diverse images, and (2) evaluate that the model is generating images that are composed of patches from different training images. For (1) we use the diversity measure introduced in SinGAN \cite{singan}, which computes the averaged standard deviation over all pixel values along the channel axis of 200 generated images. For (2) we calculate the entropy of the training images patches' probability in the generated image. Simply generating diverse images from the training set, without mixing between them, will lead to 0 entropy. 

Tab.~\ref{tab:quantitative-comparison-realism} reports these results. FewGAN is clearly superior in every qualitative metric (KID \cite{kid}, FID \cite{fid}). The margin is even larger according to the results of the user survey (realism measure). We also generate more diverse images, as can be seen by the diversity measure.
% In the diversity quantitative comparison, we achieve similar results, although higher diversity is not necessarily better, since the baseline generations are very noisy and unrealistic.

%We assess the performance of our model in terms of KID \cite{kid}. We further compute diversity measure as used in SinGAN \cite{singan}. This measure is computed as the averaged standard deviation over all pixel values along the channel axis of 200 generated images.

\begin{table}[t]
    \centering
    \begin{tabular}{@{}l@{~}c@{~~}c@{~}c@{~}c@{~}c@{}}
        \hline
         Method & KID$\downarrow$ & FID$\downarrow$ & Realism$\uparrow$ & Diversity$\uparrow$ \\
         %& \scriptsize{$\times 10^2$} & \\
         \hline
         \textbf{Our FewGAN} & 0.036 & 164 & 3.6 & 0.46\\
         HP-VAE-GAN & 0.070 & 208 & 1.7 & 0.42 \\
         Concat-SinGAN & 0.38 & 359 & 1.1 & 0.33
    \end{tabular}
    \caption{Quantitative comparison over datasets of 2,\dots,10 landscape images.}
    \label{tab:quantitative-comparison-realism}
\end{table}

% \begin{table}[H]
%     \centering
%     \begin{tabular}{c|c|c}
%         \hline
%          Method & Diversity $\uparrow$ & Entropy $\uparrow$ \\
%          \hline
%          \textbf{Ours} 
%          HP-VAE-GAN & 0.42 & 1.4 \\
%     \end{tabular}
%     \caption{\textbf{Diversity quantitative comparison} showing the Diversity and Entropy over datasets of 2,\dots,10 landscape images.}
%     \label{tab:quantitative-comparison-diversity}
% \end{table}

% \subsection{Qualitative Evaluation}
% \label{ssec:qualitative-eval}
Qualitative results can be seen in Fig. 1 for multiple landscape datasets ranging between 3 to 6 training images only. For each of these datasets, we compare unconditional generation against HP-VAE-GAN \cite{hpvaegan}. For conditional generation, we also compare against GPNN \cite{gpnn}. In addition, we demonstrate the results of our model for a variety of image manipulation tasks: editing, inpainting and harmonization. 
% Additional datasets for different number of images are given in the supplementary material. \textcolor{red}{todo: figure out if there is supp material}

% The results in this section are based on a dataset with 6 training samples, which can be seen in figure~\ref{fig:6imgs}. 

% Figure~\ref{fig:unconditional} shows a comparison of unconditional image generation between our model and HP-VAE-GAN \cite{hpvaegan}. Figure~\ref{fig:conditional} shows a comparison of conditional image generation between our model, HP-VAE-GAN \cite{hpvaegan} and GPNN \cite{gpnn}. Figures~\ref{fig:editing},~\ref{fig:inpainting} and~\ref{fig:harmonization} show the results of our model for a variety of image manipulation tasks: editing, inpainting and harmonization respectively.

\section{Conclusions}
\label{sec:discussion}

We introduced FewGAN, the first method that enables unsupervised generation of coherent images from the joint patch distribution of a small number of $N>1$ samples. We demonstrated its ability to go beyond textures and generate diverse realistic samples for natural complex images. Since this method is fully unsupervised, it has some limitations: (1) It may create a semantically wrong mapping between patches if they are nearest neighbors (e.g. map water to sky). (2) It may select different patches from different images to represent a single semantic entity (e.g. use the sky in two images to represent "sky"). Although we addressed these issues in our method - by adding positional encoding and continuity loss - they may still occur, but are less likely than with other methods we compared it with, as can be seen from our user survey and the qualitative comparison. FewGAN can provide a very powerful tool for a wide range of image manipulation tasks, as demonstrated in the qualitative evaluation.

\clearpage

\includepdf{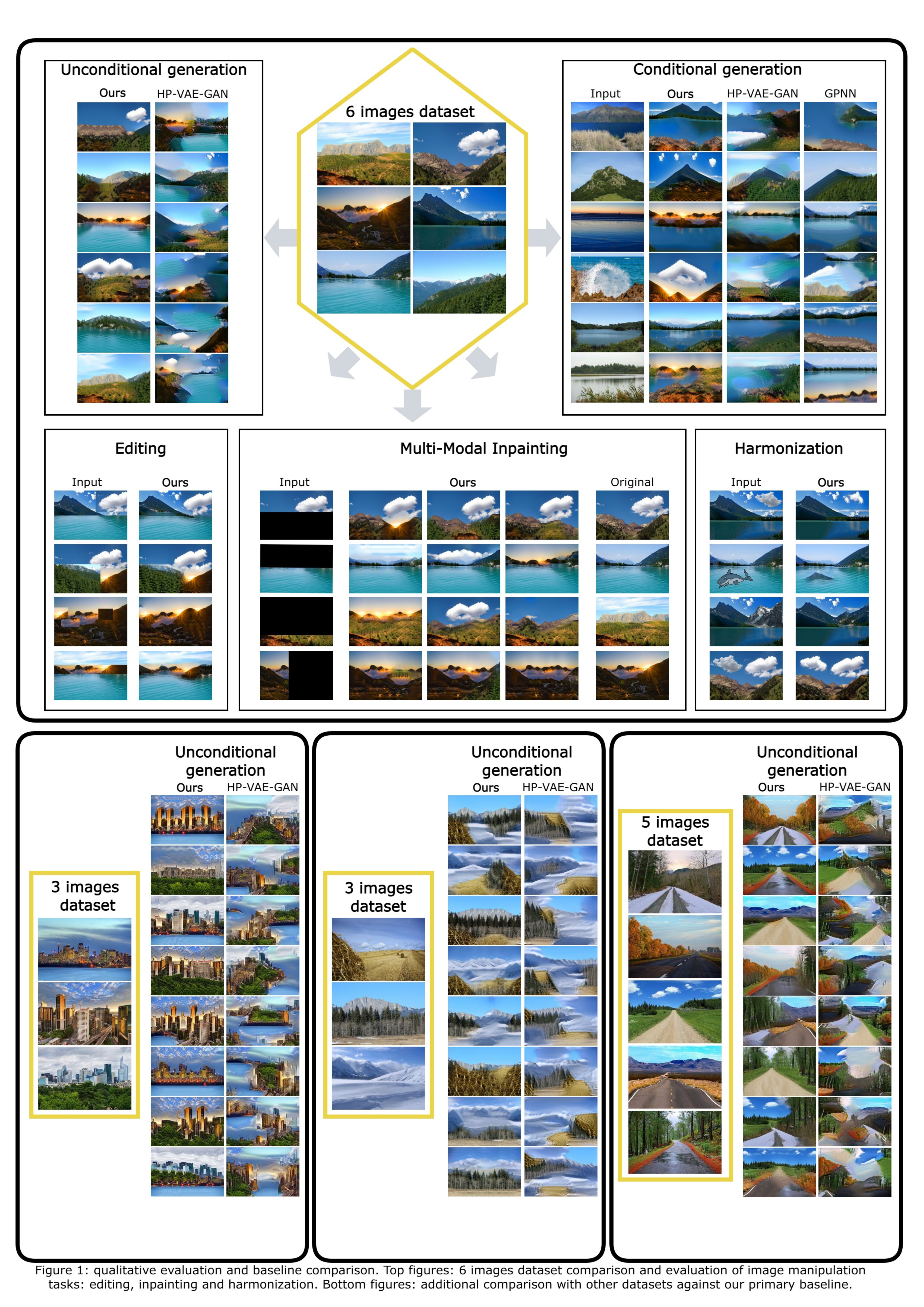}

% Below is an example of how to insert images. Delete the ``\vspace'' line,
% uncomment the preceding line ``\centerline...'' and replace ``imageX.ps''
% with a suitable PostScript file name.
% -------------------------------------------------------------------------

% To start a new column (but not a new page) and help balance the last-page
% column length use \vfill\pagebreak.
% -------------------------------------------------------------------------
%\vfill
%\pagebreak

% References should be produced using the bibtex program from suitable
% BiBTeX files (here: strings, refs, manuals). The IEEEbib.bst bibliography
% style file from IEEE produces unsorted bibliography list.
% -------------------------------------------------------------------------
\section*{Acknowledgements}
This project has received funding from the European Research Council (ERC) under the European Unions Horizon 2020 research and innovation programme (grant ERC CoG 725974).

\bibliographystyle{IEEEbib}
\bibliography{strings}

\end{document}